\documentclass[letterpaper, 10 pt, conference]{ieeeconf} 
\IEEEoverridecommandlockouts
\usepackage{cite}
\usepackage{amsmath,amssymb,amsfonts}
\usepackage{algorithmic}
\usepackage{graphicx}
\usepackage{textcomp}
\usepackage{xcolor}
\usepackage{url}

\usepackage{booktabs} \newcommand{\ra}[1]{\renewcommand{\arraystretch}{#1}}
\usepackage{caption}
\usepackage{siunitx}
\def\BibTeX{{\rm B\kern-.05em{\sc i\kern-.025em b}\kern-.08em
    T\kern-.1667em\lower.7ex\hbox{E}\kern-.125emX}}

\title{\LARGE \bf Decentralized Deep Reinforcement Learning for a Distributed and Adaptive Locomotion Controller of a Hexapod Robot
}

\author{Malte Schilling$^{1}$, Kai Konen$^{1}$, Frank W. Ohl$^{2}$, and Timo Korthals$^{3}$
\thanks{$^{1}$Malte Schilling and Kai Konen are with the Neuroinformatics Group, Bielefeld University, 33501 Bielefeld, Germany
        {\tt\small mschilli@techfak.uni-bielefeld.de}}%
\thanks{$^{2}$Frank W. Ohl is with the Department of Systems Physiology of Learning, Leibniz Institute for Neurobiology and with the Institute of Biology, Otto-von-Guericke University, Magdeburg, Germany.}
\thanks{$^{3}$Timo Korthals is with the Kognitronik and Sensorik Group, Bielefeld University, 33501 Bielefeld, Germany.}%
}

\begin{document}

\maketitle

\begin{abstract}
Locomotion is a prime example for adaptive behavior in animals and biological control principles have inspired control architectures for legged robots. While machine learning has been successfully applied to many tasks in recent years, Deep Reinforcement Learning approaches still appear to struggle when applied to real world robots in continuous control tasks and in particular do not appear as robust solutions that can handle uncertainties well. Therefore, there is a new interest in incorporating biological principles into such learning architectures. While inducing a hierarchical organization as found in motor control has shown already some success, we here propose a decentralized organization as found in insect motor control for coordination of different legs. A decentralized and distributed architecture is introduced on a simulated hexapod robot and the details of the controller are learned through Deep Reinforcement Learning. We first show that such a concurrent local structure is able to learn better walking behavior. Secondly, that the simpler organization is learned faster compared to holistic approaches.
\end{abstract}


\section{Introduction}
\begin{figure*}[tbh]
	\centering
         \includegraphics[width=\textwidth]{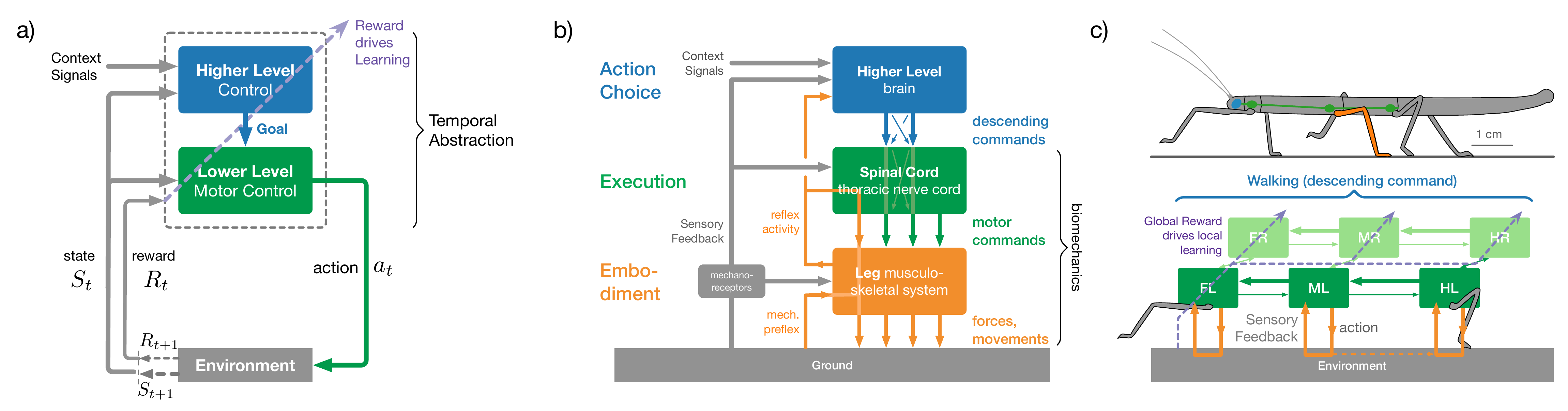}
         \caption{Visualization of influences for the biological inspired approach: On the left (a) the standard view of interaction with the environment in reinforcement learning \cite{Sutton1998} is extended to a hierarchical perspective \cite{kulkarni_hierarchical_2016} as advocated, for example in \cite{merel_hierarchical_2019}. For higher level control (shown in blue) this is in agreement with what we know on the structure of motor control in mammals \cite{arber_connecting_2018} about descending pathways and modulation of lower level control centers (shown in green) in the spinal cord (see schematic b) in blue and green). Such structures are shared not only in mammals, but also in invertebrates and insects \cite{dickinson_how_2000}, see c). Work in such simpler model systems allows more detailed analysis of interaction with the environment which has stressed the importance of very fast and local reflex activity controlled directly on the lowest level or that are even realized by passive properties as muscle elasticities or preflexes (shown in orange). One important characteristic emphasized by this work is the emergence of behavior as a result of decentralized and locally interacting concurrent control structures (bottom part of c), one example is given by the decentralized control structure found in stick insects \cite{schilling_walknet_2013}, but this concurrency is as well assumed in primates \cite{graziano_organization_2006}. In the presented approach, this decentralized architecture (bottom part of c) is used on a hexapod and learning of the six local control modules (realized as deep neural networks) is driven by a reward signal as in reinforcement learning. }
         \label{fig_biological_architectures}
\end{figure*}

Over the last years, Deep Reinforcement Learning (DRL) has been established as an exploratory learning approach (Fig. \ref{fig_biological_architectures} a) that produces effective control solutions in diverse tasks \cite{arulkumaran_brief_2017}. 
Currently, the mainstream of DRL research is addressing how to improve Deep Reinforcement Learning and in particular how to make learning more (sample) efficient as well as stabilize training \cite{nachum_data-efficient_2018}. The initial success of DRL in playing computer games \cite{mnih_human-level_2015} has shaped the field and even though there have been extensions to continuous control domains and applications in many areas \cite{arulkumaran_brief_2017}, the field is still widely dominated by approaches that deal with simulation environments. In the domain of robotics, transfer to real-world problems has proven to be difficult \cite{hwangbo_learning_2019} as it appears that the nature of such problems is fundamentally different from those in playing computer games. In most cases, simulation is first used as a tool to produce viable controllers that are only later fine-tuned, for example, on a real robot for a specific task. 

It has been argued \cite{neftci_reinforcement_2019,merel_hierarchical_2019} that some of the problems encountered in applications of DRL to real-world scenarios might be indicative of some in fact rather fundamental aspects of DRL approaches. In particular, two connected problems were identified: First, DRL is geared towards a reward and it is the explicit goal to exploit the reward structure which leads to overfitting \cite{lanctot_unified_2017}. Secondly, real world application face much more noise compared to simulations which questions the underlying assumption of a stationary Markov Decision Process \cite{kurach_google_2019} (or this becomes non-stationary as the environment or other agents are changing themselves). At its core Deep Reinforcement Learning has proven to be able to find viable solutions. But as a disadvantage it has shown that DRL is trying to exploit a reward structure as good as possible and doesn't take into account robustness of the behavior. Therefore, DRL tends to find narrow solutions that can be optimal only for a small and quite specific niche, instead of providing adaptive behavior in the sense that a control approach should be rewarded that is widely applicable in variations of a task \cite{schilling_crystallized_2019}.

As these problems have hindered application of DRL approaches in robotics when dealing with unpredictable tasks, there is now a growing interest in motor control principles in biological systems and the exploitation of these principles to improve robust adaptive behaviors in robots. For example, Merel et al. \cite{merel_hierarchical_2019} propose several such bio-inspired principles and advocate their implementation also into robot architectures. As one example they emphasize the hierarchical structure of motor control (Fig. \ref{fig_biological_architectures} b). Such hierarchies lend themselves to suitable factorization of all available information to different subsystems and to a coordinated implementation of different levels of abstraction or different timescales on which subsystems operate. It should be noted that while such approaches enfold their beneficial effects by allowing a flexible reuse and transfer of trained competences between distinct subtasks \cite{kulkarni_hierarchical_2016}  and hence improve flexibility between different (but still highly defined) contexts  \cite{frans_meta_2017,peng_deeploco:_2017}, they still do not provide the robustness in any given context that is typically found in animals (see Fig. \ref{fig_biological_architectures} a) for a visualization of hierarchical DRL).

Therefore, we here focus on another aspect of bio-inpired mechanisms already mentioned in \cite{merel_hierarchical_2019}: while higher-levels modulate the lower-level systems, these are still \textit{partially autonomous} which means that these lower-level systems can act autonomously --- for example, realizing a fast reflex pathway (see Fig. \ref{fig_biological_architectures} b) for a schematic view on biological motor control). This has been further pointed out by Sejnowski who explicitly highlighted distributed control as part of a multi-level hierarchy \cite{Sejnowski201907373}.
Such a local and decentralized control structure is in agreement with what is known on the organization of motor control in animals in general \cite{dickinson_how_2000} and in particular for locomotion in insects \cite{schilling_walknet_2013} (Fig. \ref{fig_biological_architectures} b) and c). In this paper, we want to introduce such a decentralized control principle into a control architecture for six-legged walking and want to analyze how learning in general and DRL --- as a contribution of this publication ---  benefits from this structure and how in the end control performance is affected by such a structure. Therefore, we train and compare a decentralized control architecture and compare this to learning of a baseline centralized control approach. As a result, the decentralized controller learns faster and, importantly, is not only able to produce comparable walking behavior, but even results in significantly better performance and is able to generalize as well towards novel environments. Decentralization appears as a viable principle for DRL and an important aspect for adaptive behavior \cite{merel_hierarchical_2019} as demonstrated here in our simulation on a six-legged robot.

\section{Related Work}
As DRL still struggles in high-dimensional problems and when dealing with noisy as well as uncertain environments, most of such work has been applied in simulated environments. One recent example of DRL for locomotion on a real robot was provided by Hwangbo et al. \cite{hwangbo_learning_2019}. In their approach, a policy network for a quadruped robot was initially trained in simulation. In a second step, the pretrained control network was transferred to the real robot system and further adapted on that robot. They pointed out as well a tendency of their system to overfit \cite{hwangbo_learning_2019} which is a common problem in DRL \cite{lanctot_unified_2017}. As a solution to alleviate this problem they propose a hierarchical structure \cite{merel_hierarchical_2019}.

In simulation, different approaches to locomotion already applied hierarchical structure \cite{frans_meta_2017,peng_deeploco:_2017,heess_learning_2016}. In such cases, two different levels of control were distinguished and operated on different time scales which realizes a form of temporal abstraction (Fig. \ref{fig_biological_architectures} a). Such hierarchical approaches allowed to flexibly switch between distinct subtasks and behaviors which allows to deal with a variety of distinct contexts \cite{frans_meta_2017,peng_deeploco:_2017}. For example, hierarchical DRL showed well suited for adjusting to severe interventions as the loss of a leg or when switching between obstacle avoidance, following a wall or straight walking \cite{heess_learning_2016}. This was successful for solving problems that can be solved through sequences of basic, individually learned actions, as for example a navigational task in which control should be learned from basic movement primitives. Transfer is in such cases realized as a reuse of basic motor primitives and the hierarchical organization is understood in the sense of switching between different motor primitives.

But this still does not show adaptivity within a certain behavior as found in animals which can handle broad variations within a specific context, as in climbing through a twig that only provides sparse footholds and moves unpredictably \cite{schilling_walknet_2013}. Furthermore, these simulated hierarchical approaches only dealt with a small number of degrees of freedom. One exemption is given by an evolutionary approach by Cully et al. who proposed a similar hierarchical structure and explicitly introduced an intermediate behavioral space as a lower dimensional representation for behavioral switching \cite{cully_robots_2015}. But again, this approach of switching between different behaviors depending on the current context was tested when loosing a single leg. While this poses a difficult problem and is a severe intervention, it is still a singular event that in the end required a drastic change or switch of behavior \cite{schilling_crystallized_2019}. In contrast, we want to focus on adaptivity as the way how animals deal with perturbations and variability in one given context and when using one behavior. 

Here, we take inspiration from biology and in particular research on insect locomotion. This work questions the wide held assumption that for walking there are distinct different behaviors or gaits \cite{deangelis_manifold_2019,bidaye_six-legged_2018}. Instead, there appears to be a continuum of gaits that emerge from the interaction with the environment. Such emergent behavior allows to constantly adapt to unpredictable environments and adjust the temporal coordination as required. A decentralized control structure appears to be crucial and beneficial for this kind of adaptive locomotion (Fig. \ref{fig_biological_architectures} c) gives a schematic overview of the walking controller; for details see \cite{schilling_walknet_2013}): There is one local controller for each of the six legs and the overall behavior emerges from the interaction of these six control modules. Importantly, these control modules are interacting: on the one hand, part of these interaction is directly mediated through the body as actions of a leg affect the movement of the whole animal which can be locally sensed by the other legs \cite{brooks_intelligence_1991}. On the other hand, there are local coordination rules between neighboring legs through which neighboring controllers can influence the behavior of each other \cite{schilling_walknet_2013}. 
As such a decentralized organizational structure of motor control in insects is well described \cite{durr_behaviour-based_2004,bidaye_six-legged_2018}, we will use this basic organization for our control architecture that consists of six local modules, one for each leg. Such a control structure has already been realized successfully on a six-legged robot and showed adaptivity in particular on a short timescale and when dealing with perturbations in one given context. Robust behavior emerged in such approaches that dealt well with unpredictability and small disturbances \cite{schilling_walknet_2013}. But as these controllers were handcrafted and took detailed inspiration from biology, it appears difficult to scale these towards more difficult problems as is walking through uneven terrain or climbing. Therefore, we will in this article start from such a decentralized organization in a learning based approach. This will allow to scale up to more difficult tasks and later-on to transfer to a real robot.

\section{Methods}
This study analyzes learning in architectures for a neural network based control system of a six-legged robot. First, the robot platform and simulation will be described. Second, the employed DRL framework will be referenced. Third, as the main contribution of this article, the decentralized control architecture and a baseline structure will be introduced.

\subsection{PhantomX Robot Simulation}
The PhantomX MK-II is the second generation of a six-legged robot distributed by Trossen Robotics \cite{phantomx2020}. It consists of six legs each actuated by three joints. The main body consists of a rectangular plexiglas base (\SI{260}{\milli\meter} by \SI{200}{\milli\meter}) and for each leg, the first joint is attached to the main body in a sandwich type construction (mounted between two such plexiglas sheets). Therefore, the first joints moves the leg forward and backward, while the other two joints lift respectively extend the leg. For our learning approach, joint limits were restricted to avoid limb- and leg-collisions as well as unpractical leg postures, e.g. postures in which the tibia would touch the top of the torso. 

The basic structure is loosely inspired from insects and has comparable degrees of freedom. All joints are common revolute joint servo motors (Dynamixel AX-12A). While the original PhantomX robot can be controlled using a preinstalled control system on a Arduino compatible microcontroller, it has been adapted to the Robotic Operating System (ROS) \cite{ochs2018}.

In this study, we use as a first step only a simulated robot\footnote{robot definitions: \url{https://github.com/kkonen/PhantomX}} in the ROS simulation environment \textit{Gazebo} \cite{gazebo2004} as it is open source and because of the good support of ROS (here ROS 2, version Dashing Diademata \cite{quigley2009ros}). As a physics-engine \textit{Bullet} was used \cite{coumans2015}. The simulation required several custom modules (e.g., publishing joint-effort controllers in ROS 2) that have been implemented now for ROS 2\footnote{for details see the open repository \url{https://github.com/kkonen/ros2learn}}. One current difference between simulated robot and the real robot was the implementation of ground contact sensors using data from the simulation environment, but solutions for such sensors have been proposed for the robot as well \cite{tikam2018posture}.
Simulator and ROS were run with a fixed publishing-rate of \SI{25}{\hertz} in order to align sensor and actuator data. This guaranteed a stable and constant step-size. 

\subsection{Deep Reinforcement Learning Framework}
Reinforcement Learning is characterized by an agent that is interacting in an environment (Fig. \ref{fig_biological_architectures} a). While the agent produces actions, it gets as a response an updated state of the environment and a reward signal. The goal of the agent is to learn a policy $\pi(S)$ for sequential decision making that maximizes the accumulated long-term return. During learning the agent has to explore possible alternatives and while getting more confident should come up with a policy it can exploit (for an introduction see \cite{Sutton1998,arulkumaran_brief_2017}). Sequential decision making can be formalized as a Markov Decision Process which provides and is defined as a tuple of:
\begin{itemize}
\item observable states $\mathcal{S}$, 
\item possible actions $\mathcal{A}$,
\item a reward signal $\mathcal{R}$ providing the immediate reward after choosing an action from the current state, 
\item transition probabilities $\mathcal{P}$ that describe the probability distribution over states after following an action when being in the current specific state, and
\item a discount factor $\gamma$ that describes how to decrease the weights of future rewards. 
\end{itemize}

We are dealing with a continuous control task: the state space as given by the sensory signals is continuous as well as the action space which corresponds to the joint motor signals. Overall, these spaces are high dimensional and continuous which requires to rely on function approximation for learning a probabilistic policy. Deep neural networks are used in Deep Reinforcement Learning as function approximators that map observations to actions \cite{arulkumaran_brief_2017}. 

As a framework for DRL we employed OpenAI's gym. The connection to the simulator through ROS was wrapped as a new environment through a connector using \textit{ROS2Learn} \cite{ros2learn}. This allowed to use the standard baseline implementations of DRL algorithms \cite{openai2017baselines}. In this approach Proximal Policy Optimization (PPO) \cite{schulman2017ppo} was used as it has shown to work well on continuous tasks without the need of intensive hyperparameter tuning. As a further advantage, it has a comparably low sampling complexity (amount of training samples required to find usable policies).

The reward function was defined as the mean velocity over an episode (1024 iteration steps, equals nearly $41$ seconds). The agent uses the received reward $R_{t+1}$ after executing an action $A_t$ in state $S_t$ in order to directly optimize its' policy $\pi$. A neural network was used as a function approximator for the learned policy. The policy networks consisted of two hidden layers of each $64$ units with $tanh$ activation functions. Updating the neural networks' weights is called an epoch. After training, the policy network should generalize to novel sensed states and modulate motor output in order to achieve a maximal external reward. The next section will describe in detail the two compared architectures that employ such policy networks.

\begin{figure}[tb]
	\centering
         \includegraphics[width=\columnwidth]{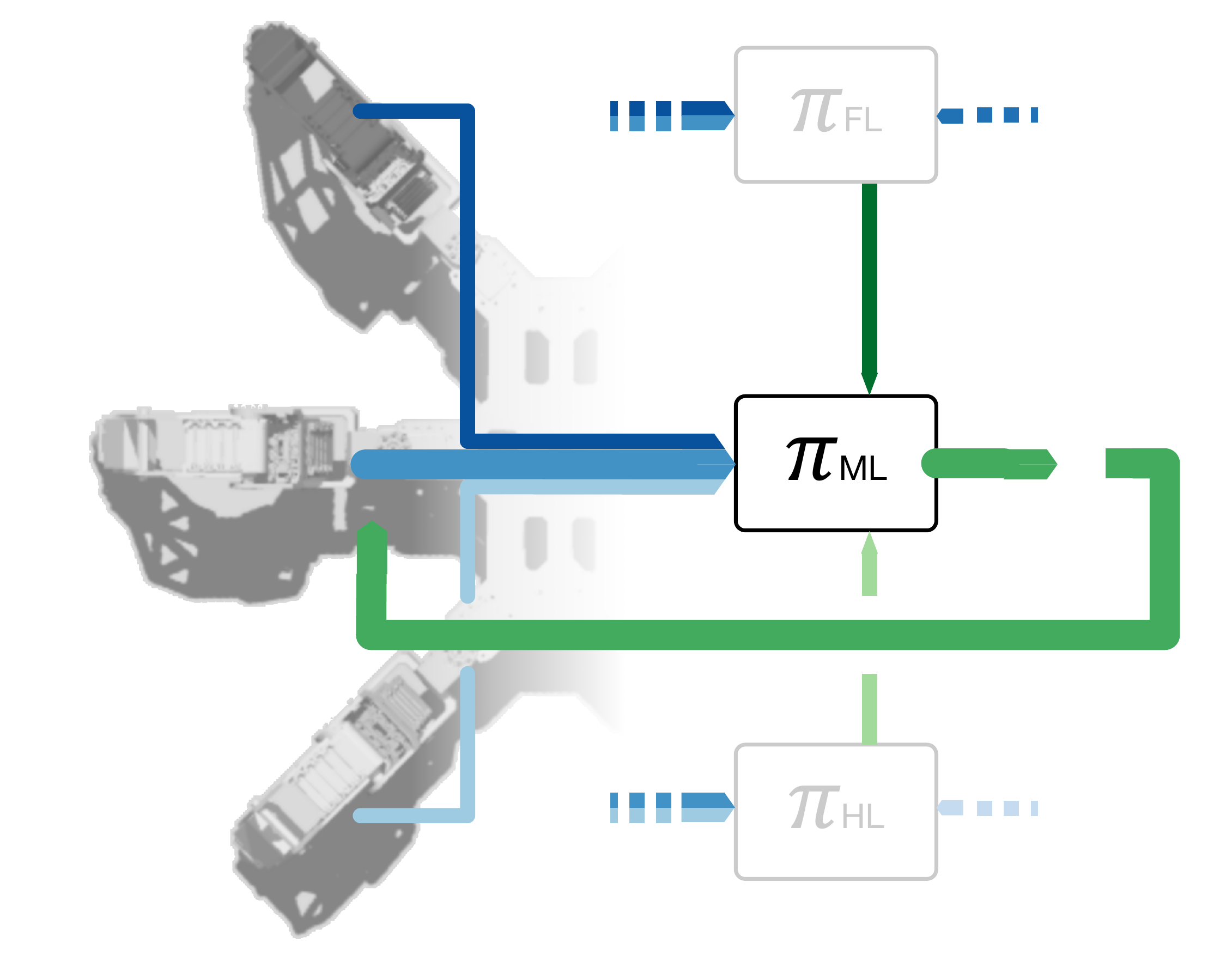}
         \caption{Overview of the decentralized control architecture, shown for the middle left leg (ML): inputs to the controller are sensory information (blue) from the leg and its' two neighbors and last actions of the neighboring controllers to provide context. Control output (joint activations for only the respective leg) are shown in green. For each leg there is an individual controller in the decentralized structure.}
         \label{fig_decentralized}
\end{figure}

\subsection{Decentralized Control Architecture}
The main contribution of this paper is the decentralized control architecture. Importantly, the presented approach is not hierarchically structured using a higher level and switching between different behaviors (Fig. \ref{fig_biological_architectures} shown in a) in blue), but realizes the decentralized structure on the lower control level (Fig. \ref{fig_biological_architectures} c) shown in green). This decentralized structure is embodied in a six-legged robot. There is one local control module for each leg. Each controller consists of a single policy neural network (two hidden layers with $64$ hidden units in each of these layers). Input to each of the leg controllers is only local information stemming from that particular leg and the two neighboring legs (Fig. \ref{fig_biological_architectures} c) shows these influences as green arrows). Furthermore, there is interaction with the environment that can mediate information between legs (Fig. \ref{fig_biological_architectures} c) orange arrows). The observation space of one controller (Fig. \ref{fig_decentralized} shows as an example the middle left leg controller and its input as well as output space) is $42$ dimensional as it receives information from the controlled leg (positions of the three leg segments and ground contact). In addition, a local controller gets information from the two neighboring legs (including positions of leg segments, ground contact, and last action signal from the neighboring legs). Last, local controllers get information on the orientation of the body as a six-dimensional vector \cite{zhou2018d6}.

\begin{figure}[b]
	\centering
         \includegraphics[width=\columnwidth]{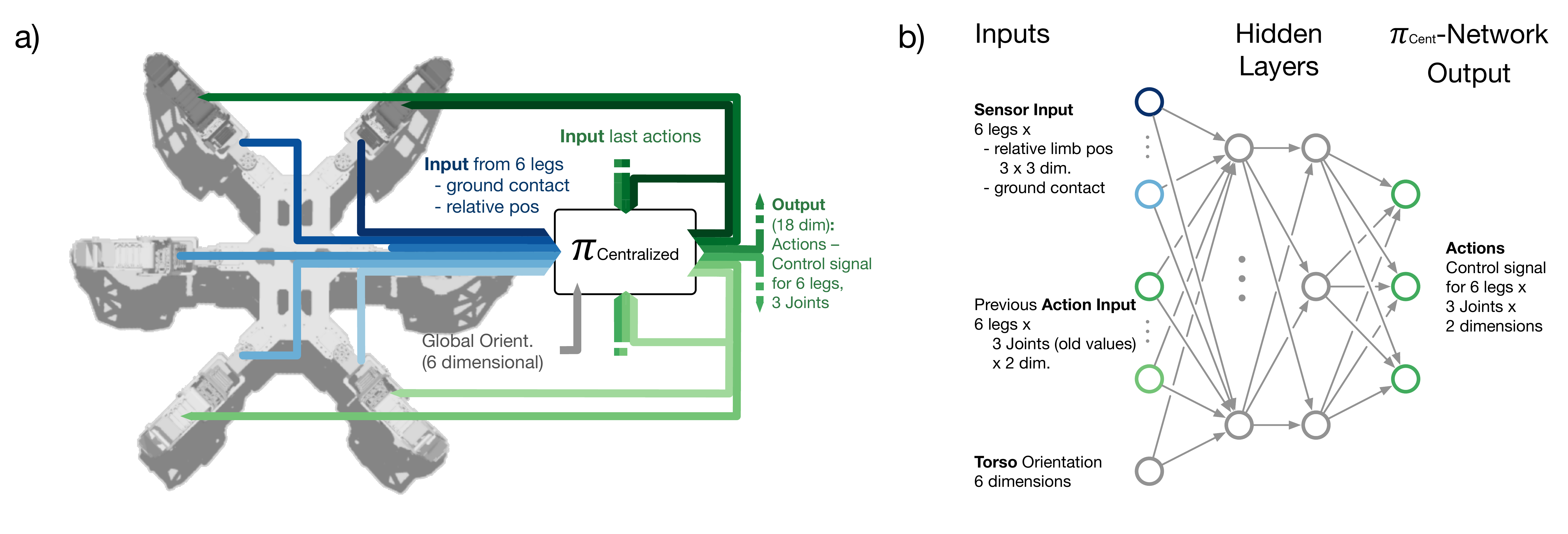}
         \caption{a) Overview of the centralized architecture that is trained as a single-agent network. All sensory information from the legs (blue) are fed into the policy network which produces joint actions for all $18$ joints (shown in green). b) View of centralized policy network: a neural network with two hidden layers (64 hidden units in each hidden layer). Sensory input is given as information from leg and output are joint actions (for the decentralized approach: six such networks are used which each produce joint activations for a single leg and only receives partial sensory inputs).}
         \label{fig_centralized}
\end{figure}

During Reinforcement Learning the six controllers learn in parallel which is comparable to multi-agent learning. But in our decentralized architecture, the individual controllers are part of the same robot and can have access to information from the other control agents. In contrast to a fully centralized approach, this information is restricted in our approach and only neighboring legs provide information.

The decentralized architecture is compared to a centralized (single-agent) DRL approach: for this baseline, a single policy network is used (Fig. \ref{fig_centralized}). All available information is used as an input (overall $84$ dimensions) and the task for the policy network is to learn optimal actions for all $18$ joints. 

As the centralized approach has access to all information, the input space and space of possible decisions is high dimensional. It appears more difficult to explore and find an optimal policy. In contrast, for the decentralized approach the input and output space of each of the six local control modules is lower dimensional. While this simplifies exploration, it is not clear if all necessary information is available to the local controllers. Our experiments address, first, if the information provided to the local control modules is sufficient to learn competent walking behavior. And, secondly, how the lower dimensional input and control space affect learning and if there is faster learning observable.

\section{Results}
We hypothesize, first, that a decentralized architecture consisting of six local leg controllers can produce stable locomotion behavior. Secondly, that a local structure simplifies learning. As a last aspect, we want to analyze how such a control structure can generalize and copes with novel environments. In the following, the learning experiments will be shown and analyzed that compare a centralized structure with the proposed decentralized structure.\footnote{Simulation environment is made available as a docker. Instructions and all data from the experiments can be found at: \url{https://github.com/malteschilling/ddrl_hexapod}}

\begin{figure}[tb]
	\centering
         \includegraphics[width=\columnwidth]{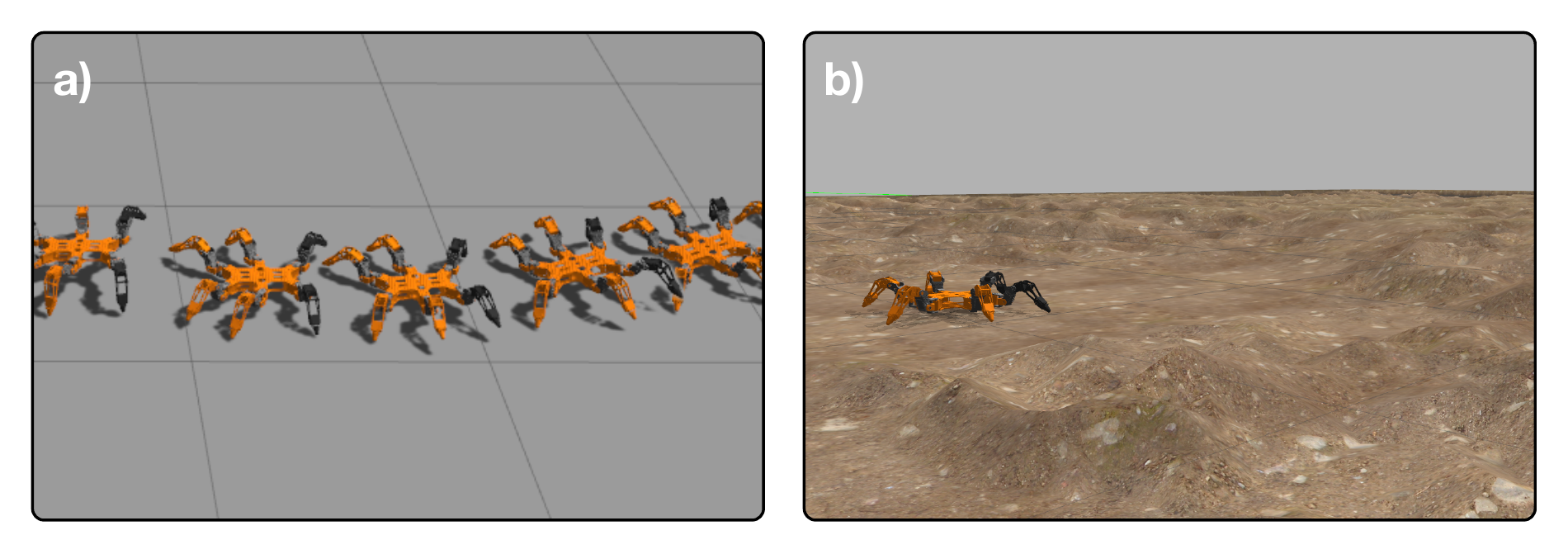}
         \caption{Visualization of the simulated PhantomX robot. a) shows a walking sequence on flat terrain. b) shows the uneven terrain condition in comparison to the robot.}
         \label{fig_walking_beh}
\end{figure}

\subsection{Performance of Locomotion Architectures on Flat Terrain}
As a first experiment, we studied learning to walk on a flat terrain and compared as the two conditions the two different architectures. The controllers were trained through DRL from scratch. Each of the two architectures was trained $15$ times using different random seeds for $5000$ epochs.

Both learning approaches were able to learn walking behavior (see Fig. \ref{fig_walking_beh} a) and supplemental video) with quite a high velocity that looked well coordinated. The trained controllers from the last epoch were afterwards evaluated. For each controller for over $100$ individual episodes the performance (mean velocity over the episode) was collected. Individual results from the different seeds are given in the supplement (see data repository). For the centralized baseline approach the mean performance was $546.70$ (standard deviation $131.23$). The proposed decentralized architecture consisting of six local control modules reached a performance of $657.11$ (std.dev. $68.07$). First, we wanted to test that the decentralized architecture does not perform worse. Therefore, we performed a two-tailed Welch t-test (following \cite{colas_hitchhikers_2019} with the null hypothesis that the two approaches perform on a similar level). The null hypothesis could be rejected (p-value of $.011$) and we can conclude that the decentralized approach performed significantly better compared to the centralized approach. The mean rewards were actually substantially higher for the decentralized approach with a relative effect size of $1.02$ which indicates a large effect size.

The results showed a large variance which is typical for DRL-based solutions. In many cases, the focus is therefore on the best solutions produced by running multiple seeds during learning. Therefore, we compared as well the best performing controllers. Overall, the standard deviation in both conditions were in a similar range, but the distribution of the centralized approaches appeared larger. In table \ref{tab_best_seeds_flat} the ten best ranked controllers are shown. As can be seen from these results, the best performing approach was a centralized approach, but there seems to be an upper limit for maximum mean velocity and multiple approaches converged towards this performance. Overall, we found more decentralized architectures in the best performing approaches which was confirmed by a rank-sum test showing that the decentralized architectures performed significantly better (Wilcoxon–Mann–Whitney test, $p=.041$).

\begin{table}[tbh]
\centering
\ra{1.1}
\captionsetup{format =hang}%

\caption{Detailed results after learning for $5000$ epochs. The mean rewards were evaluated for each of the fifteen learned controllers for the two different architectures in 100 simulation runs. Note that standard deviation is taken for the single controller over the $100$ repetitions which showed to be low. }

\begin{tabular}{@{}p{0.1cm}rlcccr@{}}
\toprule
& Rank & Architecture & Seed & Avg. Reward & Std. Dev. &
\\ \midrule
& 1. & Central/Baseline & 2 & $770.158$ & $21.52$ \\
& 2. & Decentralized & 1 & $764.209$ & $26.01$ \\
& 3. & Decentralized & 5 & $756.096$ & $27.50$\\
& 4. & Decentralized	& 7 & $733.194$ & $56.82$\\
& 5. & Decentralized & 12 & $721.685$ & $37.84$\\
& 6. & Decentralized	& 2 & $707.283$ & $33.84$\\
& 7. & Decentralized & 6 & $695.860$ & $24.94$\\
& 8. & Decentralized & 4 & $681.352$ & $26.10$\\
& 9. & Central/Baseline & 5 & $652.852$ & $18.28$\\
& 10. & Decentralized & 14 & $640.512$ & $30.23$\\
\bottomrule
\end{tabular}
\label{tab_best_seeds_flat}
\end{table}

To summarize: As a first result, we found that a decentralized control architecture is able to produce high performance and well coordinated walking behavior. The local controller even showed significantly better performance with a large effect size.

\subsection{Comparison of Learning}
\begin{figure}[tb]
	\centering
         \includegraphics[width=\columnwidth]{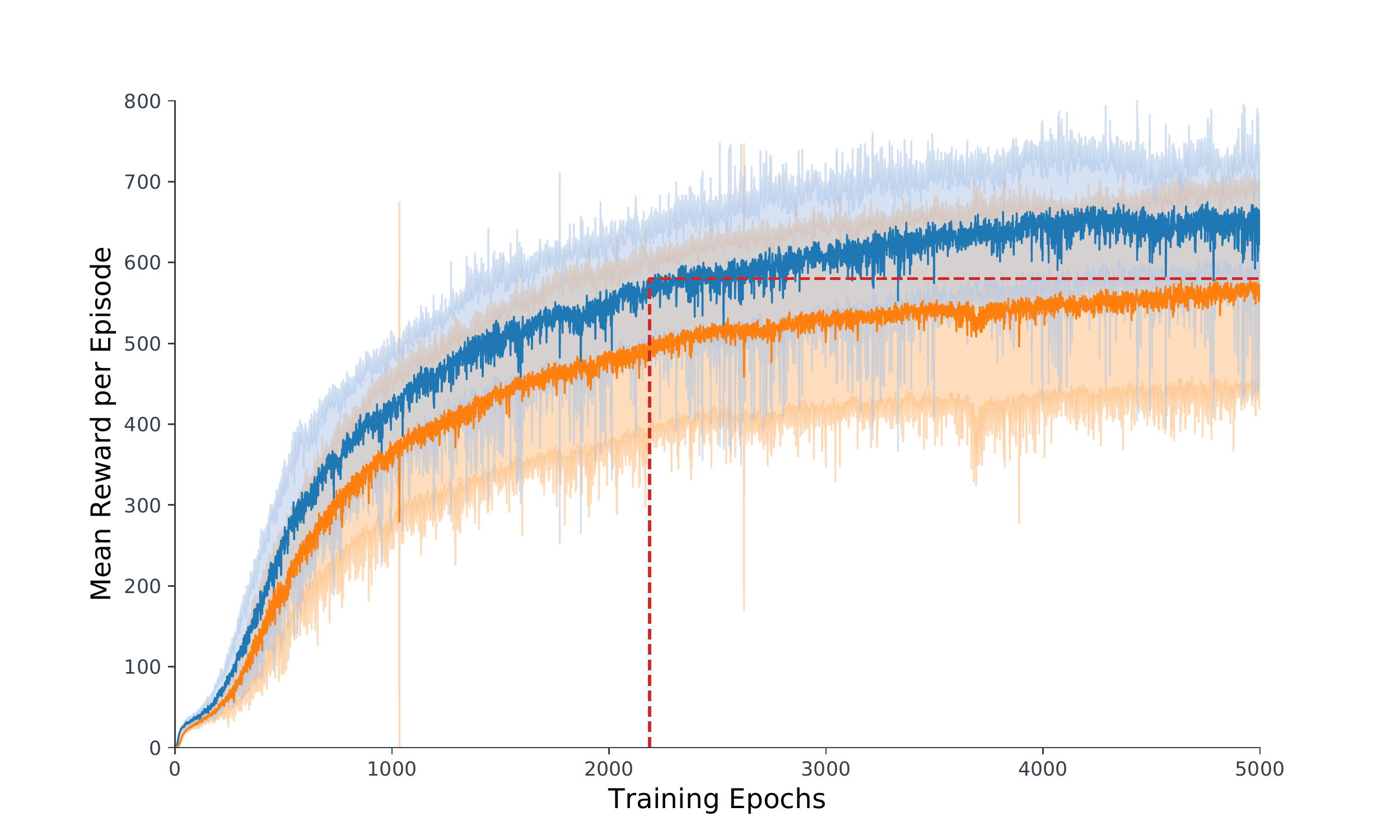}
         \caption{Comparison of the mean reward during training: mean performance over the fifteen decentralized controllers is shown in blue and mean reward for the baseline centralized approach is shown in orange (shaded areas show standard deviation). Performance is measured as reward per episode. Seeds were only measured up to 5000 epochs points as learning appears to have converged by then. The horizontal red dashed line visualizes the maximum of the centralized approach at the end of training. The vertical red dashed line indicates when this performance level was reached by the decentralized approach (after $2187$ epochs already).}
         \label{fig_mean_flat}
\end{figure}

The learned architectures were trained for $5000$ epochs as initial tests showed that at this point controller performance converged for both cases (running simulations for up to $12000$ epochs only showed minor further improvements which appear to confirm the reasonable idea of a maximum reachable velocity). In this section, we will look at the development of training performance over time (Fig. \ref{fig_mean_flat}). 

Mean performance was calculated over the fifteen seeds for each of the two architectures over training. Looking at individual runs showed a high variance during learning. While the mean calculation smoothed the reward signal, this is still visible. Such jumps in performance are typical for DRL approaches. Learning progressed in several stages (details not shown) from simple crawling towards walking lifted from the ground. Comparing the two different architectures, the difference in performance between these is clearly visible. The decentralized architecture reached a higher reward level earlier and appears to learn much faster. Compared to the mean reward of the centralized approach at the end of training, the decentralized approaches reached such a level already after $2187$ epochs. 

\begin{table*}[tbh]
\centering
\ra{1.1}
\captionsetup{format =hang}%

\caption{Comparison of rewards between the different control architectures during evaluation. Data was collected during evaluation for $100$ episodes for each controller. Given are mean rewards (and standard deviation) for each group of controllers (each group consisted of fifteen individually trained controllers). A Welch t-test was performed to compare the two conditions (decentralized and centralized architecture) with the p-value given in the last column. 
}

\begin{tabular}{@{}p{0.1cm}lccccccr@{}}
\toprule
\multicolumn{2}{l}{Condition} &  \multicolumn{2}{c}{Decentralized Arch.} & \phantom{abc}&  \multicolumn{2}{c}{Centralized Arch.} &\phantom{abc}& p-value\\  \cmidrule{3-4} \cmidrule{6-7}
&& Mean Reward & std.dev. && Mean Reward & std.dev. \\
\midrule
\multicolumn{2}{l}{Evaluation on flat terrain}\\
& Trained on flat terrain & $657.11$ & $68.07$ && $546.70$ & $131.23$ && $.011$\\
& Generalization, trained on height map & $492.69$ & $79.22$ && $441.50$ & $48.00$ && $.050$\\
\\
\multicolumn{2}{l}{Evaluation on uneven terrain}\\
& Generalization, trained on flat terrain & $397.91$ & $83.87$ && $334.06$ & $112.94$ && $.102$\\
& Trained on height map & $424.30$ & $40.09$ && $382.42$ & $51.23$ && $.023$\\
\bottomrule
\end{tabular}
\label{tab_generalization}
\end{table*}

\subsection{Generalization in Transfer to Uneven Terrain}
\begin{figure}[tb]
	\centering
         \includegraphics[width=\columnwidth]{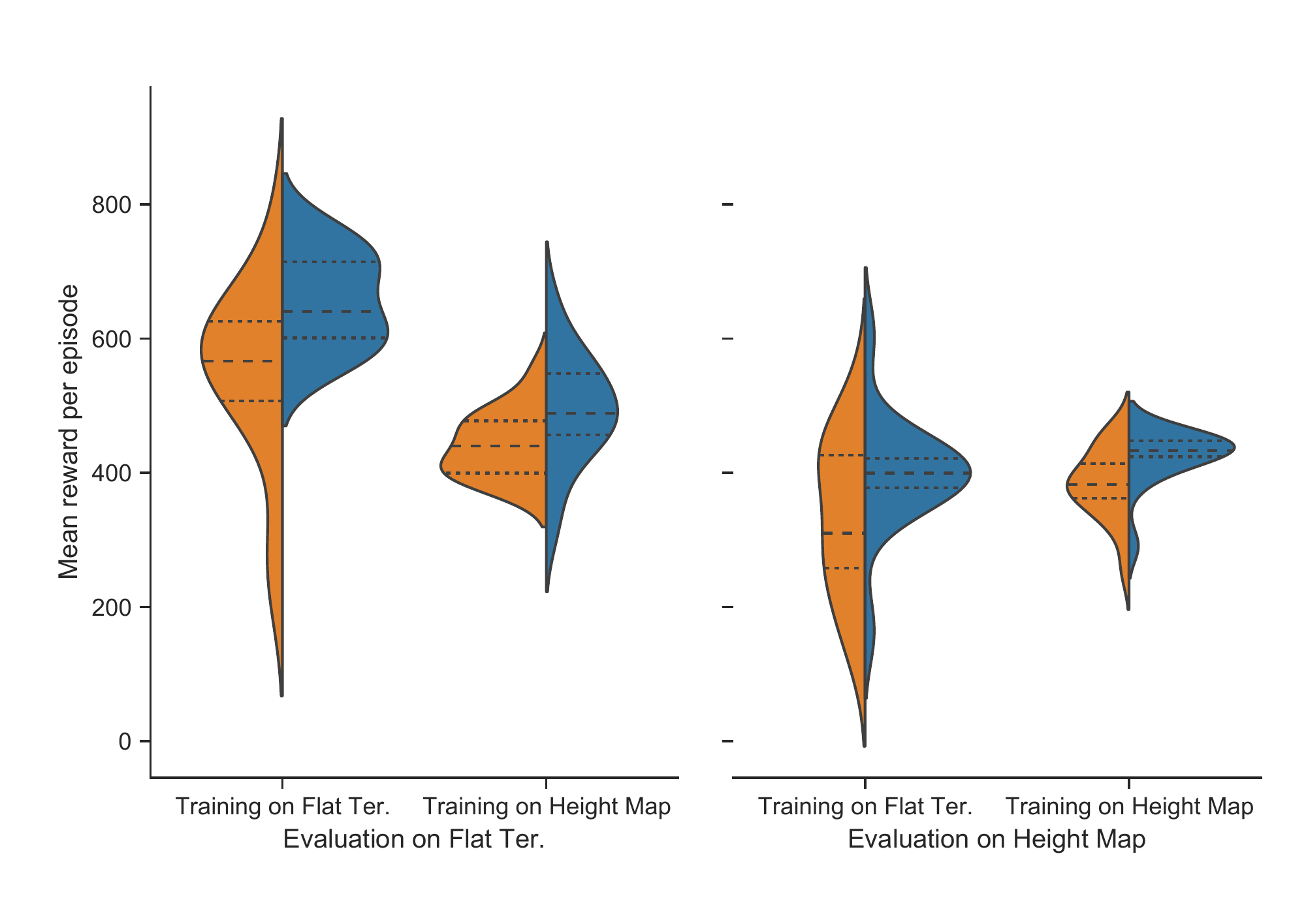}
         \caption{Comparison of the (approximated) performance distribution for the controllers. Shown are violin plots for two evaluation conditions: left shows the evaluation of walking on flat terrain and right shows walking on uneven terrain (modulated through a height map with maximal height of \SI{0.10}{\m}). Inside the two evaluations, it is further distinguished between controllers trained on the flat terrain (first and third column) and controllers trained on the height map (second and fourth column). The violin plots show to approximated distributions (and the quartiles are given inside): orange shows the centralized baseline approach and blue shows the decentralized architecture. Performance for the fifteen different seeds for each condition was measured as mean reward per episode (indicates mean velocity over simulation time).}
         \label{fig_terrain_comp}
\end{figure}

In order to understand how such a controller deals with unpredictability, we further evaluated the controller for a novel, uneven terrain it hadn't experienced during learning. All controllers trained on the flat terrain were again tested for $100$ simulation runs each on uneven terrain. The terrain was generated from height-maps with a maximum height of \SI{0.10}{\m} (Fig. \ref{fig_walking_beh} b), further variations can be found in the data repository). Height-maps were generated using the diamond-square algorithm \cite{miller1986diamond}. Detailed results are given in table \ref{tab_generalization} and see Fig. \ref{fig_terrain_comp}. 

First, there was an expected drop-off in performance for the uneven terrain for both architectures (Fig. \ref{fig_terrain_comp} shown between left most results and results in the third column that shows distribution of performance on the height map). We further trained fifteen additional controllers for both architectures on the uneven terrain. When comparing results on the uneven terrain, there was no significant difference between controllers that had to generalize (as they were trained on flat terrain) and the specialized controllers that were trained on the same uneven terrain (Fig. \ref{fig_terrain_comp} two rightmost columns, comparison of the two different training conditions evaluated on the uneven terrain). Again, we found a significant difference for the expert condition in which the controllers trained on the height map and tested on the height map were compared: in this case, the two-tailed Welch t-test again showed a significant difference ($p=.020$) for mean performance in favor of the decentralized architecture which confirms our earlier results. Inside the generalization condition, difference in performance of the two architectures was reduced to a trend ($p=.102$).

Last, the controllers trained on uneven terrain were evaluated on the flat terrain. Here, the controllers trained on the uneven terrain don't reach the level of the specialist controllers that were trained on the flat terrain and there is a significant performance difference. This was true for decentralized and centralized approaches to the same extent, but again the decentralized architecture performed better. The performance of the controllers was only slightly above their performance on uneven terrain.

To summarize: It appears that decentralized controllers generalize well towards the more difficult task as they perform on the same level as controllers that were explicitly trained on the task. In comparison to the baseline, decentralized controllers were on the same performance level.

\section{Discussion and Conclusion}
In this article, we presented a biologically-inspired decentralized control architecture for a hexapod robot. The architecture consists of six local control modules that each control a single leg. As each controller only has access to local information, we wanted to analyze if the local information is sufficient to produce robust walking behavior. Therefore, we tested the decentralized architecture on a six-legged robot in a dynamic simulation and analyzed the learning behavior in a DRL setup. There are two main results from our study: First, the decentralized architecture produced well performing and coordinated controllers that as a group even showed significantly better performance compared to the baseline centralized approach.

Second, the learning task for the decentralized approach appears simplified as the dimensionality (input dimensions as well as control dimensions) is drastically reduced. This lead to faster learning of well performing controllers and earlier convergence towards an assumed maximum velocity. The performance of the centralized approach was reached by decentralized architectures in less than half the training time. Furthermore, the distribution of performances appears narrower. In contrast, for the centralized approach it appears as learning got stuck more often in local minima. Decentralization therefore appears as a viable principle that helps facilitate Deep Reinforcement Learning and in continuous control tasks allow to speed up learning as well as avoid local minima.

Considering generalization towards a novel and supposedly more difficult task, both approaches showed similar good performance that was on the level of controllers trained on the novel task. From behavioral research in insects \cite{schilling_walknet_2013}, we would have assumed that decentralization would facilitate adaptive behavior and lead to better generalization capabilities \cite{schilling_crystallized_2019}. But we couldn't show such an advantage. This might be explained by the selection of the reward function which aimed only for high velocities. In fast walking in animals, it is assumed that the influence of sensory feedback is reduced in faster walking and running, and coordination is more driven by a simpler synchronization signal \cite{ijspeert_central_2008}. As future work, it is therefore important to use a wider variety of tasks during training. This is further emphasized by the fact that after training on the uneven and supposedly more difficult terrain, the controller didn't perform as good on the flat terrain. In hindsight, this appears reasonable as such controllers have converged in the more demanding terrain to a different maximum velocity that is appropriate for the uneven terrain and takes the disturbances into account. But after learning to walk on such a terrain, the controller is suddenly tasked with literally running without having to consider possible disturbances. 

As a next step, we want to extend the current analysis to further strengthen our results. Furthermore, we want to analyze the structure of the learned controllers. Zahedi et al. \cite{zahedi_higher_2010} used a local control approach for a chain of wheeled robots in a learning approach. They provided a geometric interpretation assuming that such local policies form a low-dimensional subfamily of the family of all possible policies. Our results confirm their findings as our experiments support that such a restriction to a lower-dimensional subclass appeared as not too restrictive even for the much more complicated and higher dimensional case of locomotion. Analyzing the different controllers might shed further light on the underlying geometric principles and how decentralization helps to constrain optimization in a meaningful way.

In the future, we want to extend training to more diverse and demanding tasks, similar to curriculum learning \cite{bengio_curriculum_2009}. This will allow us to analyze how decentralized approaches generalize: Does a local structure contribute to the robustness and adaptivity of behavior or does a decentralized approach tend to overfit as well? Furthermore, this will include integrating more task demands into the reward function without running the risk of bias due to reward shaping \cite{heessTSLMWTEWER17}. In particular, as our long term goal is application on the real robot, the reward function has to reflect physical properties as well, for example penalizing high torques. 

\section*{Acknowledgment}
The authors thank Chris J. Dallmann and Holk Cruse for discussions and helpful comments. Furthermore, we want to thank the Computer System Operators of the Faculty of Technology for their technical support.

\bibliographystyle{IEEEtran}
\bibliography{ReferencesZotero}

\end{document}